\documentclass[Journal,letterpaper,NoLineNumbers]{ascelike-new}

\usepackage[utf8]{inputenc}
\usepackage[T1]{fontenc}
\usepackage{lmodern}
\usepackage{graphicx}
\usepackage[figurename=Fig.,labelfont=bf,labelsep=period]{caption}
\usepackage{subcaption}
\usepackage{amsmath}

\usepackage{newtxtext,newtxmath}
\usepackage[colorlinks=true,citecolor=red,linkcolor=black]{hyperref}

\usepackage{array}
\usepackage{subcaption}
\usepackage{balance}
\usepackage{longtable}
\usepackage[T1]{fontenc}
\usepackage[flushleft]{threeparttablex}
\usepackage[symbol*]{footmisc}
\usepackage{hyperref}
\usepackage{booktabs}
\usepackage{pifont}
\usepackage{algorithm}
\usepackage{caption}
\usepackage{algpseudocode}
\usepackage{svg}
\usepackage{graphicx}
\usepackage{multirow}
\usepackage{booktabs}

\newsavebox\mybox

\NameTag{Ma et al., \today}

\let\svthefootnote\thefootnote
\newcommand\freefootnote[1]{%
  \let\thefootnote\relax%
  \footnotetext{#1}%
  \let\thefootnote\svthefootnote%
}

\begin{document}

\title{Bridging the Reality Gap in Digital Twins with Context-Aware, Physics-Guided Deep Learning}

\author[1]{Sizhe Ma}
\author[2]{Katherine A. Flanigan}
\author[3]{Mario Berg\'es}

\affil[1-3]{Civil \& Environmental Engineering, Carnegie Mellon University,
Pittsburgh, PA, USA, 15213; email: $^1$sizhem@andrew.cmu.edu; $^2$kflaniga@andrew.cmu.edu; $^3$mberges@andrew.cmu.edu}

\maketitle
\freefootnote{Mario Berg\'es holds concurrent appointments at Carnegie Mellon University (CMU) and as an Amazon Scholar. This manuscript describes work at CMU and is not associated with Amazon.}

\begin{abstract}
Digital twins (DTs) enable powerful predictive analytics, but persistent discrepancies between simulations and real systems---known as the reality gap---undermine their reliability. Coined in robotics, the term now applies to DTs, where discrepancies stem from context mismatches, cross-domain interactions, and multi-scale dynamics. Among these, context mismatch is pressing and underexplored, as DT accuracy depends on capturing operational context, often only partially observable. However, DTs have a key advantage: simulators can systematically vary contextual factors and explore scenarios difficult or impossible to observe empirically, informing inference and model alignment. While sim-to-real transfer like domain adaptation shows promise in robotics, their application to DTs poses two key challenges. First, unlike one-time policy transfers, DTs require continuous calibration across an asset's lifecycle---demanding structured information flow, timely detection of out-of-sync states, and integration of historical and new data. Second, DTs often perform inverse modeling, inferring latent states or faults from observations that may reflect multiple evolving contexts. These needs strain purely data-driven models and risk violating physical consistency. Though some approaches preserve validity via reduced-order model, most domain adaptation techniques still lack such constraints. To address this, we propose a Reality Gap Analysis (RGA) module for DTs that continuously integrates new sensor data, detects misalignments, and recalibrates DTs via a query-response framework. Our approach fuses domain-adversarial deep learning with reduced-order simulator guidance to improve context inference and preserve physical consistency. We illustrate the RGA module in a structural health monitoring case study on a steel truss bridge in Pittsburgh, PA, showing faster calibration and better real-world alignment.
\end{abstract}

\section{Introduction} \label{sec:intro}

Digital twins (DTs) have gained widespread attention as transformative platforms that integrate high-fidelity virtual models with physical assets. By continuously incorporating sensor data and domain-specific knowledge, DTs stand to enable real-time monitoring, predictive analytics, dynamic system optimization, and decision support throughout an asset’s lifecycle \cite{ma_state---art_2024}. Like any model designed to represent a real-world system, DTs encounter persistent discrepancies between simulated outputs and real-world measurements---a divergence commonly referred to as the reality gap \cite{muller_self-improving_2022}. Here, the reality gap specifically refers to errors that result when the virtual model does not fully capture real-world asset behavior. These errors are systematic, often arising from an incomplete understanding of the system, unmodeled dynamics, or computational constraints and persist until addressed through improved modeling, real-world feedback, or adaptive mechanisms. Historically, they have constrained the real-world applicability of models, particularly in robotics, where unmodeled dynamics can render learned controllers ineffective in real-world deployment \cite{salvato_crossing_2021}. These concerns are relevant to digital twins, whose predictive capabilities can degrade when their digital representation drifts from the system's physical behavior \cite{foundational_2024}.

These predictive challenges underscore the need to systematically address sources of deviation between digital models and their real-world counterparts. One challenge in bridging the reality gap is reducing \textit{context mismatch}, which arises when the physical twin's operating environment, lifecycle stage, or usage conditions no longer align with the assumptions encoded in the DT. In many DT applications, context captures external conditions and usage parameters that govern how a system functions, such as environmental factors, operational modes, and load profiles \cite{hribernik_autonomous_2021}. As a physical system progresses from design to prototyping, and then to long-term deployment and maintenance, its context evolves in ways that may invalidate the digital twin's underlying assumptions. If these contextual changes are not reflected in the digital twin, the model's predictions can systematically deviate from reality, leading to performance degradation over time. Addressing context mismatch thus requires explicit strategies to keep the DT synchronized with the physical asset's ongoing transformations.

Another obstacle in mitigating the reality gap is reducing \textit{cross-domain mismatch}, wherein a single DT often spans multiple subsystems, such as mechanical, electrical, software, or operational subsystems. An error originating in one domain (e.g., drift in an electrical sensor) can cascade into other subsystems (e.g., mechanical, operational), compounding discrepancies throughout the model \cite{Heindl2022}. This cross-domain coupling complicates error localization, as it is no longer sufficient to treat each subsystem in isolation \cite{Rasheed2020}. Ensuring consistency across domains becomes pivotal for preserving accuracy, particularly when domain-specific effects interact in unanticipated ways  \cite{ma_state_2}.

A third challenge is reducing \textit{multi-scale mismatch}, which becomes evident when a DT aims to address the reality gap across multiple temporal or spatial scales. While some predictive tasks focus on high-frequency, component-level phenomena (e.g., local vibration analyses), others require long-term strategic insights (e.g., maintenance scheduling) \cite{Yang2023_neuro}. Techniques designed for one scale may not generalize to another, leading to mismatch if the model's scope does not encompass the entire range of scales encountered in real-world operations. Consequently, bridging the reality gap in a DT demands error management strategies capable of consistent performance across disparate scales.


While cross-domain and multi-scale mismatches are well-studied in DT research, context mismatch remains a critical yet underexplored challenge. When real-world conditions deviate from design assumptions, the DT's validity can be compromised. Proper calibration to the physical system's context could greatly reduce such discrepancies \cite{kapteyn_data-driven_2022}. Many DT models rely on structural or parametric assumptions tied to operating conditions like temperature, load, and boundary constraints \cite{ma_state_2}. If these assumptions don't align with reality, predictive accuracy suffers. Yet in live settings, context is often latent or only partially observable, making direct measurement of key calibration factors infeasible \cite{wen2022digital}. As a result, existing methods must infer context mismatches from limited sensor data---e.g., temperature, strain, or accelerometer readings---which often fail to fully capture system complexity \cite{liu_domain_2020}.


Unlike real-world operational settings, simulation environments offer complete control over contextual variables, enabling a ``perfect knowledge'' scenario where all parameters are explicitly defined \cite{peng_sim--real_2018}. For instance, in structural health monitoring, one can precisely specify friction coefficients, boundary conditions, and load scenarios \cite{i3ce2025}, then generate virtual measurements---such as stress distributions or vibration signals---that closely mimic real-world observables. This stands in stark contrast to the partial, noisy, or indirect data available in operational settings. The challenge, then, is bridging these extremes: limited, uncertain real-world observations versus fully specified simulation conditions. Leveraging the strengths of both could enhance inference and DT calibration, helping to close the reality gap.

\begin{figure}
\centering
\begin{subfigure}[b]{0.9\textwidth}
   \includegraphics[width=0.9\linewidth]{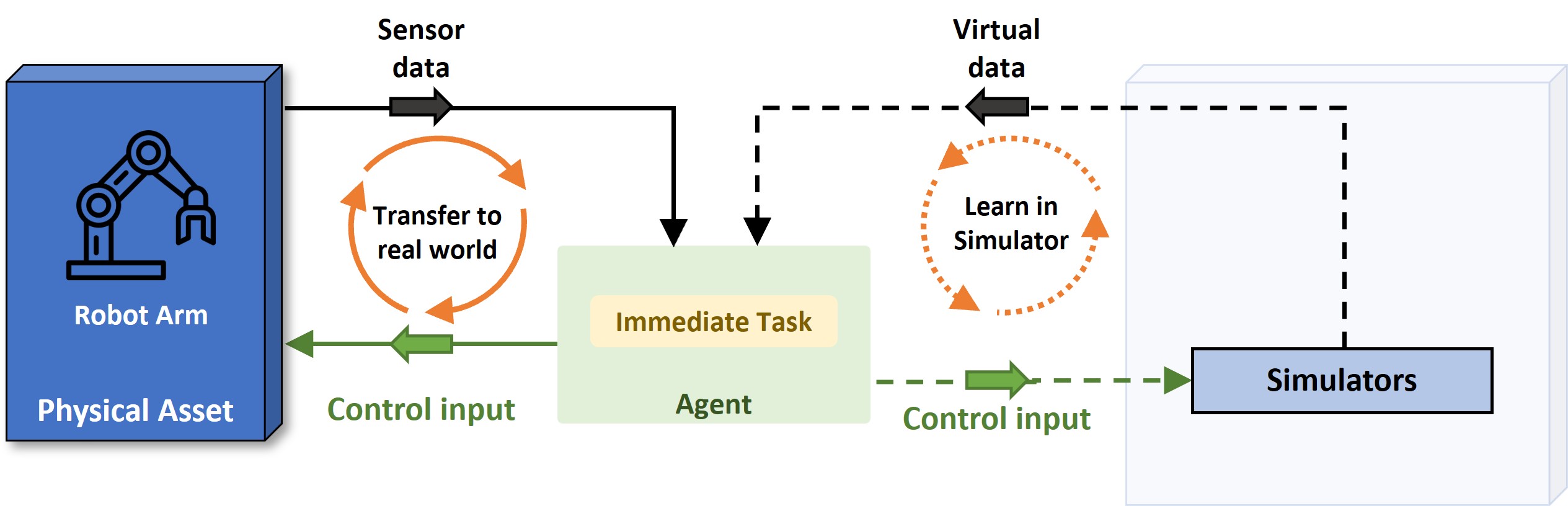}
   \caption{}
   \label{fig:sim-to-real-mechanism} 
\end{subfigure}
\hfill
\begin{subfigure}[b]{0.9\textwidth}
   \includegraphics[width=0.9\linewidth]{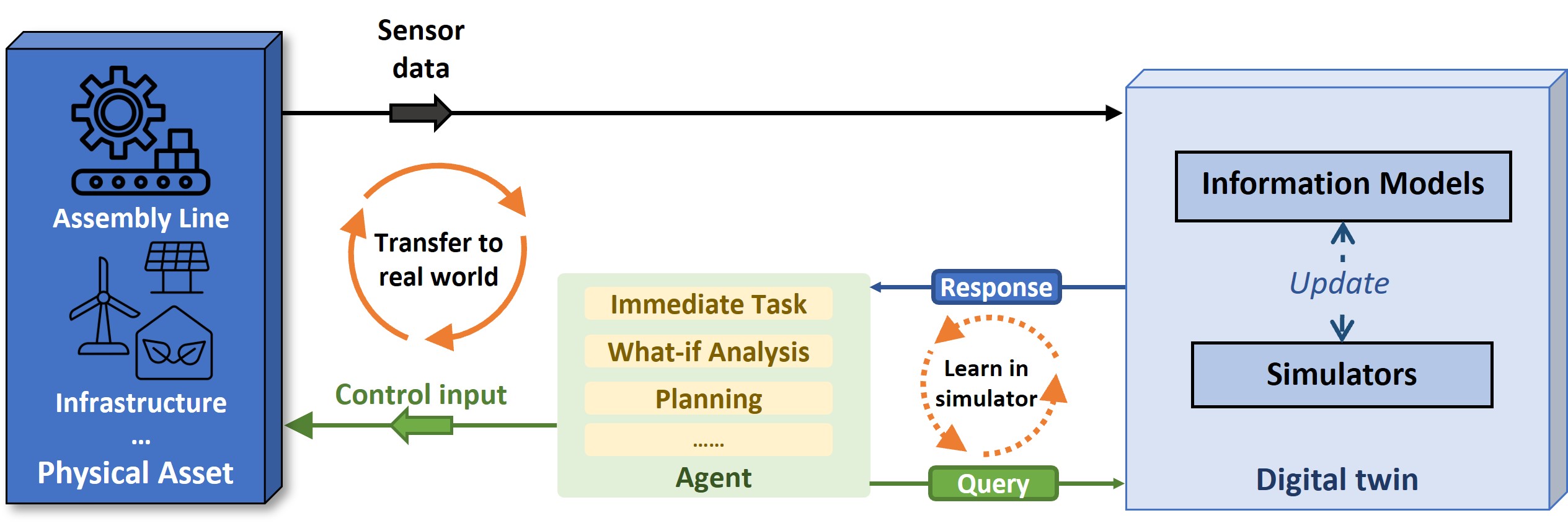}
   \caption{}
   \label{fig:digital-twin-mechanism}
\end{subfigure}

\caption{A comparison of how simulator knowledge (represented by the agent) is used in the real world for (a) robotics and (b) DT applications. In (a), knowledge learned in the simulator is transferred directly to the physical asset, as the processing pipelines for simulation and real-world deployment are typically similar---especially in task-based robotic applications. In (b), however, the DT must remain continuously synchronized with evolving operational data, creating richer opportunities for predictive modeling and what-if analysis but also introducing significant challenges because the real world's synchronized pipeline can diverge substantially from the simulator environment.}
\label{fig:sim-to-real-digital-twin-dif}
\end{figure}

One way to bridge the gap between real-world uncertainty and simulation precision comes from robotics, where sim-to-real transfer adapts control policies trained in simulators to robots \cite{salvato_crossing_2021}. This approach mitigates modeling discrepancies by adjusting learned behaviors to real conditions, but it is typically task-specific---focused on object manipulation or path following---rather than maintaining a continuously synchronized virtual system. In robotics, the goal is robust control, not broader functions like what-if analysis or long-term prediction \cite{peng_sim--real_2018}. Though recent robotics work has adopted the term ``digital twin,'' most implementations lack continuous context updates and sustained synchronization between virtual and physical systems \cite{liu_digital_2022}. As shown in Figure~\ref{fig:sim-to-real-digital-twin-dif}, sim-to-real setups (Figure~\ref{fig:sim-to-real-mechanism}) involve one-time knowledge transfer with minimal feedback, while DTs (Figure~\ref{fig:digital-twin-mechanism}) maintain ongoing synchronization through automated query-response mechanisms. This enables predictive modeling and scenario analysis but requires the DT to remain up to date with evolving system conditions. While sim-to-real techniques offer useful insights for reducing the reality gap, their task-centric focus limits direct applicability to domains like infrastructure monitoring or industrial diagnostics, where continuous context inference is essential. This distinction introduces two key challenges.


First, in robotics, a single sim-to-real transfer often suffices for mission-specific tasks \cite{zhao2020transfer}. In contrast, domains like structural health monitoring require continuously updated DTs that remain accurate over an asset's lifecycle, adapting to evolving operating conditions \cite{ritto2021digital}. A central challenge is structuring information flow across diverse components to support timely and coherent context updates. Three issues are critical: (a) \textit{Role definition}: identifying which entities (e.g., sensors, models, processing modules) exchange data and under what conditions; (b) \textit{timing and synchronization}: establishing communication frequency and managing inevitable out-of-sync updates; and (c) \textit{data integration}: incorporating new observations without losing historical context or introducing inconsistencies. Robotics pipelines typically do not address these needs, as their focus is on transferring a policy that performs acceptably, not maintaining a continuously calibrated DT for long-term monitoring and analysis.


A second key difference lies in how sensor data is used to infer internal system states. In robotics, some ambiguity is acceptable if control actions still achieve the desired outcome \cite{celemin_knowledge-_2023}. In contrast, applications like infrastructure maintenance or failure risk assessment demand much higher precision, as a single observation may correspond to multiple possible contextual states. This ambiguity complicates downstream tasks---such as what-if analyses or scheduling---that rely on accurate state inference \cite{kessels_real-time_2023}. Moreover, sim-to-real methods in robotics rarely enforce domain-specific physical constraints, like material fatigue, energy balances, or conservation laws \cite{zhao2020transfer}. For DTs, however, maintaining such constraints is essential for predictive accuracy and reliability. Developing methods that resolve state ambiguity while embedding physical principles remains an open challenge, highlighting a key gap between task-focused robotics and the broader demands of DT applications.

Seen in Figure~\ref{fig:sim-to-real-digital-twin-dif}, sim-to-real transfer relies on one-way knowledge transfer from a simulator to a physical asset, without maintaining continuous updates. In contrast, applications like manufacturing or infrastructure management require synchronization between the DT and the real world to support context inference and long-term predictive modeling. To address these expanded requirements, we propose a \emph{Reality Gap Analysis (RGA)} module with an explicit query-and-response mechanism, embedded within a modular DT architecture. Our approach extends sim-to-real techniques beyond robotics by incorporating simulation-driven knowledge to infer and update the digital twin’s context, thereby reducing the reality gap. Crucially, the RGA continually recalibrates as the physical system evolves, maintaining robust alignment between the virtual model and its real-world counterpart. 

To address the first challenge, we explicitly define each component's role, establish structured communication protocols, and specify when and how new data integrate into the existing digital twin. These design elements ensure that context changes are captured systematically, maintaining DT accuracy throughout the asset’s lifecycle. For the second challenge, the RGA incorporates a deep learning framework with domain adaptation, a technique used in robotics to align simulated and real-world data distributions. Unlike traditional sim-to-real transfer, our approach enforces domain-specific physical constraints by integrating a reduced-order simulator into the inference process. This combination improves predictive accuracy, stabilizes convergence, and enhances interpretability, supporting the use of DTs for a broader range of analytic and predictive tasks. 

We illustrate the effectiveness of our approach through a case study on condition-based monitoring for a bridge at Carnegie Mellon University (Pittsburgh, PA), demonstrating how the RGA module enhances precise knowledge transfer from simulation to operational environments by dynamically updating the DT based on observed sensor data while enforcing physics-based constraints. This demonstration also highlights the digital twin's improved decision-support capabilities, particularly for what-if analysis, where stakeholders can explore the potential effects of hypothetical stressors on bridge performance.  By continuously refining its virtual representation in response to real-world conditions, our approach ensures that the DT remains an effective tool for predictive maintenance, structural risk assessment, and lifecycle planning.

\section{Limitations of Calibration and Domain Adaptation Techniques} \label{sec:lr}

To motivate our proposed approach, this section reviews prior work on two foundational challenges for DT fidelity: (1) the need for continuous, context-aware calibration as operating conditions evolve; and (2) the use of domain adaptation techniques that preserve physical consistency and interpretability. For each area, we synthesize existing methods, highlight their limitations in the context of digital twins, and clarify the specific gaps our framework addresses.

\subsection{Continuous Context Capture for Calibration}
DT applications such as structural health monitoring, energy systems, and industrial diagnostics require sustained synchronization between virtual and physical representations to support long-term predictive tasks. While sim-to-real transfer methods in robotics have inspired progress in this area, they are primarily designed for one-off policy transfers in mission-specific tasks \cite{peng_sim--real_2018}, lacking the continuous calibration needed in DTs \cite{ma2025standard}.

Efforts to enable continuous calibration have focused on three key challenges: role definition, synchronization timing, and data integration. First, the question of role definition---determining which system components exchange information and under what conditions---has been explored through agent-based, ontology-based, and inference-based approaches. Agent-based methods conceptualize the twin as an autonomous decision maker that adapts based on incoming data \cite{vrabic_intelligent_2021}, while ontology-based frameworks define roles using semantic structures to facilitate interoperability and context sharing \cite{bao_ontology-based_2022}. Inference-based approaches 
prioritize data-driven adaptation, relying on statistical learning to detect drift and recalibrate \cite{song_online_2022}. All three categories have limitations: agent-based systems require predefined utility metrics, ontologies demand expert-crafted domain models, and inference-based approaches often struggle with inverse problems under unfamiliar conditions. In short, existing role definitions remain heavily reliant on static prior knowledge, limiting their capacity for autonomous, data-driven adaptation.

The second challenge is timing and synchronization---deciding when the DT should update to stay aligned with the physical system. Synchronization strategies typically fall into four categories: time-driven, event-driven, hybrid, and adaptive \cite{alghamdi2024synchronization}. Time-driven methods offer predictability but can waste resources during periods of stasis \cite{jia_digital-twin-enabled_2021}, while event-driven updates conserve bandwidth but may miss slow-evolving changes if triggers are not well defined \cite{lopez_real-time_2021}. Hybrid approaches combine the two, increasing system complexity and risk of race conditions \cite{gehrmann_digital_2020}. Adaptive synchronization strategies show promise by adjusting update frequencies based on observed conditions \cite{han_dynamic_2023}, but often rely on heuristic thresholds that are difficult to tune and interpret \cite{tan_digital_2024}. These limitations point to the need for a self-adjusting, data-informed synchronization policy that responds to system changes in a principled way.

The third challenge is data integration---incorporating new observations without compromising historical context or introducing conflicts. Dual-model frameworks preserve a historical baseline while updating a data-driven surrogate \cite{molinaro_embedding_2021}, and library-based methods store predefined component models for reuse \cite{kapteyn_data-driven_2022}. While both approaches support model continuity, they lack systematic mechanisms for detecting and integrating truly novel contexts. In particular, they offer limited guidance on when to expand the historical repository or how to validate the distinctiveness and credibility of new data.

Collectively, these limitations highlight the need for a more adaptive and data-driven approach to DT calibration—one that can structure inter-component roles without fixed ontologies, trigger recalibration responsively based on observed system behavior, and selectively integrate novel context in a principled way. Meeting these requirements calls for a framework that treats calibration not as a one-time process, but as a continuous, context-aware dialogue between the physical and digital systems. In what follows in Section \ref{sec:method}, we describe how our proposed RGA module operationalizes this vision, enabling robust, interpretable, and lifecycle-spanning calibration through real-time sensor feedback and simulation-guided inference.

\subsection{Physically Constrained Inference and Deep Learning-Based Domain Adaptation}

Domain adaptation is a critical technique for sim-to-real transfer, particularly in robotics where aligning simulated and real-world feature distributions is key to robust control. Approaches range from feature-level alignment to pixel-space transformations. For instance, \cite{tzeng2014deepdomainconfusionmaximizing} introduced pairwise constraints to align simulation and real-world visual embeddings, while \cite{gupta_learning_2017} developed invariant feature spaces to transfer skills learned in simulation to robots. Generative adversarial networks have also been used to minimize visual domain gaps at the pixel level, as demonstrated by \cite{bousmalis_unsupervised_2017}. These methods, alongside others reviewed in \cite{zhao2020transfer}, have shown that reducing domain discrepancies---either through adversarial learning, statistical distance minimization, or image translation---is essential for improving transferability.

Recently, DT research has begun adopting similar domain adaptation strategies to narrow the simulation-to-reality divide. Two dominant approaches have emerged. Discrepancy-based methods minimize predefined statistical divergences, such as maximum mean discrepancy, between simulation and real-world feature distributions \cite{zhang2023digital}. Adversarial approaches, on the other hand, employ a domain discriminator to enforce domain-invariant feature learning through a minimax training process \cite{liu_domain_2020}. The flexibility of adversarial learning makes it especially attractive for digital twins, where hand-crafting alignment criteria may be impractical due to high-dimensional, nonlinear dynamics.

However, adapting these techniques directly to DTs presents unique challenges. First, unlike many robotics tasks that prioritize control performance, DTs are often used for inverse modeling: inferring latent states or fault conditions from observable data. This inverse problem is ill-posed, as multiple contextual states may produce similar measurements \cite{kessels_real-time_2023}. Second, their operating conditions often evolve over time (i.e., domain gap is not static). Without mechanisms for continual adaptation or recalibration, well-aligned models may quickly become outdated \cite{foundational_2024,salvato_crossing_2021}. Third, deep domain adaptation models---especially adversarial ones---are notoriously difficult to train, often requiring finely tuned hyperparameters and careful balancing of loss terms \cite{ganin_domain-adversarial_2016}.

Another key limitation is that most domain adaptation models in DT research lack integration with physical knowledge. This omission limits interpretability and can lead to spurious mappings that violate known physical relationships. Physics-informed neural networks offer one solution by embedding governing equations into the training process \cite{raissi_physics-informed_2019}, but they require explicit, closed-form representations of the physics---an assumption often violated in digital twins, where high-fidelity simulators encapsulate the governing dynamics implicitly \cite{ma_state---art_2024}. To bridge this gap, recent work has explored reduced-order models as a more tractable alternative. For example, \cite{kim_fast_2022} use a pre-trained reduced-order model to compute physically grounded loss signals within a deep learning pipeline, thereby enforcing physical plausibility without requiring full equation-based formulations.

Our approach incorporates a pre-trained reduced-order model directly into the domain adaptation training loop as a physics-guided constraint. This design helps to mitigate the inverse modeling challenge by ensuring that inferred contextual states not only align statistically with real-world data but also produce outputs that are physically consistent. In tandem with a continuously calibrated DT framework, our method dynamically monitors and updates the model when the domain gap widens. While this increases the complexity of the training objective, it also enhances stability and accuracy by providing robust physical regularization and improved gradient flow.

\section{Reality Gap Analysis Module Design and Implementation}\label{sec:method}

The RGA module is a structured approach for continuously calibrating the DT as operational conditions change. This section outlines our modular framework, including its core components, data exchange mechanisms, and adaptation strategies. We begin with the \textit{DT architecture}, which integrates real-world sensor data with simulation models. Next, we present the RGA module, enabling \textit{continuous context calibration} through structured information flow, out-of-sync state detection, and adaptive data integration. Finally, we introduce a \textit{deep learning-based inference model} that applies domain adaptation while enforcing physical consistency via a reduced-order simulator, ensuring DT updates remain both data-driven and physically valid.

\subsection{DT Architecture and Foundations for Reality Gap Analysis}

Before addressing the challenges of reality gap mitigation, it is essential to establish a DT architecture that supports continuous synchronization with the physical asset. This section introduces the foundational DT framework upon which the RGA module operates, detailing its modular structure, data exchange mechanisms, and role in preserving model fidelity over time.

\begin{figure}
    \centering
    \includegraphics[width=1\linewidth]{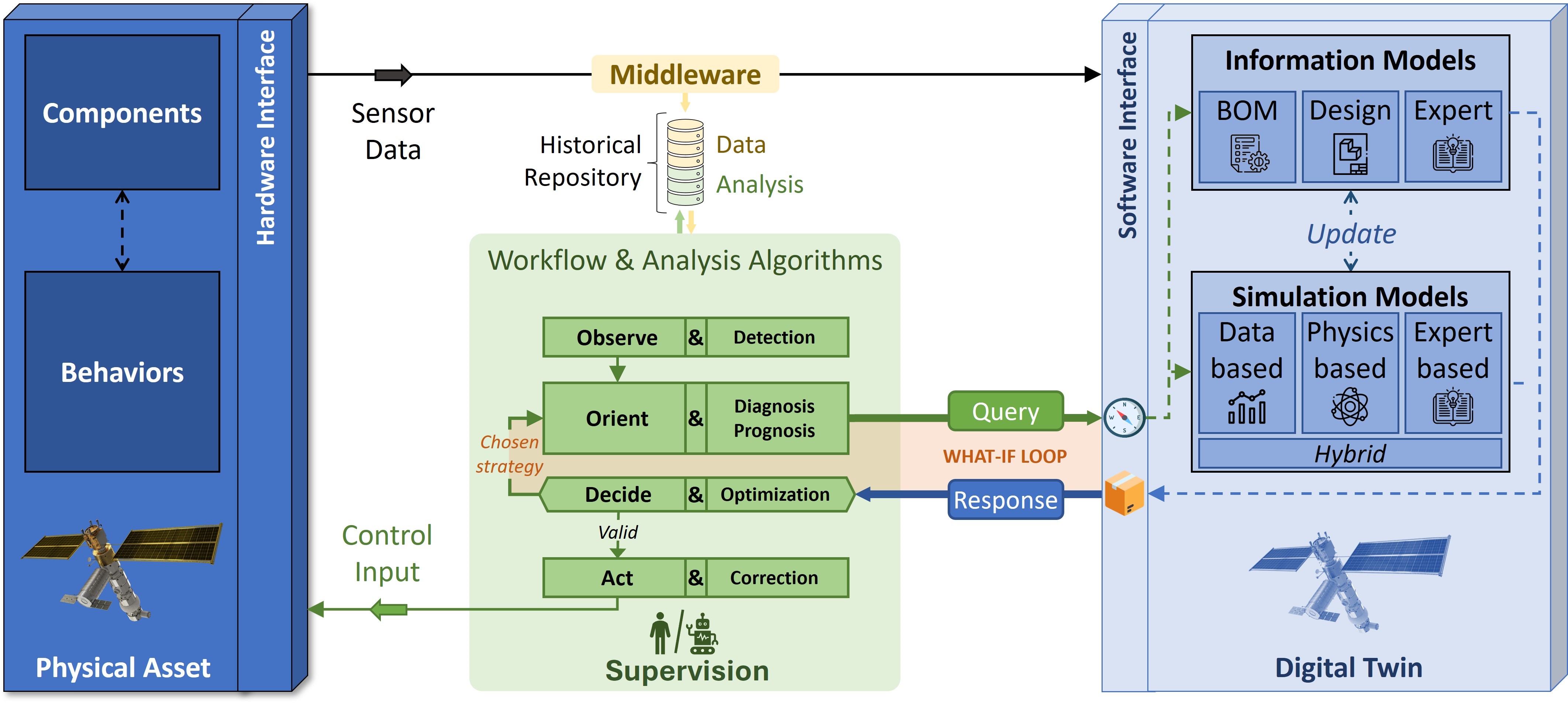}
    \caption{DT framework from our earlier work \protect \cite{ma2025framework}.}
    \label{fig:framework_original}
\end{figure}

Building on our prior work on DT architectures, we adopt a unified, modular framework designed to address the two key challenges introduced earlier in Section~\ref{sec:intro} \cite{ma2025framework}. A visual overview of this DT architecture is provided in Figure~\ref{fig:framework_original}. This framework enables real-time data flow between the physical asset and its virtual counterpart, supporting continuous updates and adaptive calibration. The core data pipeline functions as follows: (1) The \textit{physical asset} continuously transmits sensor data to the \textit{hardware interface}, which streams these measurements to the \textit{middleware}. (2) The middleware serves as a coordination hub, distributing sensor data to the \textit{digital twin} and \textit{historical repository}, which stores time-indexed records for long-term archival and retrieval. (3) The DT consists of two primary components: \textit{information models} that encode semantic metadata such as geometry, material properties, and relational schemas, and \textit{simulation models} that generate forward-looking predictions based on physics-informed, data-driven, or expert-derived techniques. These models are updated in near real-time via the middleware, ensuring continuous, bidirectional synchronization with the physical asset.

A key feature of this architecture is its functional separation between the DT and \textit{workflow and analysis algorithms}. The DT serves as a stable, unified virtual proxy of the physical system that can be accessed or queried by any external decision-making process. This decoupling is maintained through a \textit{query and response} module that acts as a controlled interface. Query and response ensures that downstream tasks---such as what-if analysis, diagnostics, and predictive maintenance---can operate without directly altering the DT. It also ensures that the integrity and consistency of the DT remains intact, preventing ad hoc modifications that could introduce inconsistencies.

Several elements of this framework are critical for mitigating the reality gap. First, the historical repository archives both sensor and simulation data across the asset's lifecycle, enabling comparative analysis of evolving conditions under operational contexts. Second, the query and response mechanism acts as an intermediary for controlled calibration events, facilitating structured communication between the DT and analytical processes. Together, these elements provide a stable yet adaptable foundation for ensuring that the DT remains accurate and responsive to real-world changes. This modular, data-driven architecture forms the basis for the RGA module, which systematically identifies and corrects deviations between the digital and physical systems.

\subsection{Reality Gap Analysis for Continuous Calibration} \label{sec:gap1}

Maintaining an accurate DT requires continuous adaptation to evolving real-world conditions. Without a structured approach to integrating new sensor data, detecting model drift, and preserving historical knowledge, the DT may diverge from reality. To address this, we introduce the RGA module, which supports continuous calibration through a systematic query-and-response mechanism. This section details the information flow, drift detection strategies, and data integration techniques that underpin our approach.

\begin{figure}
    \centering
    \includegraphics[width=0.9\linewidth]{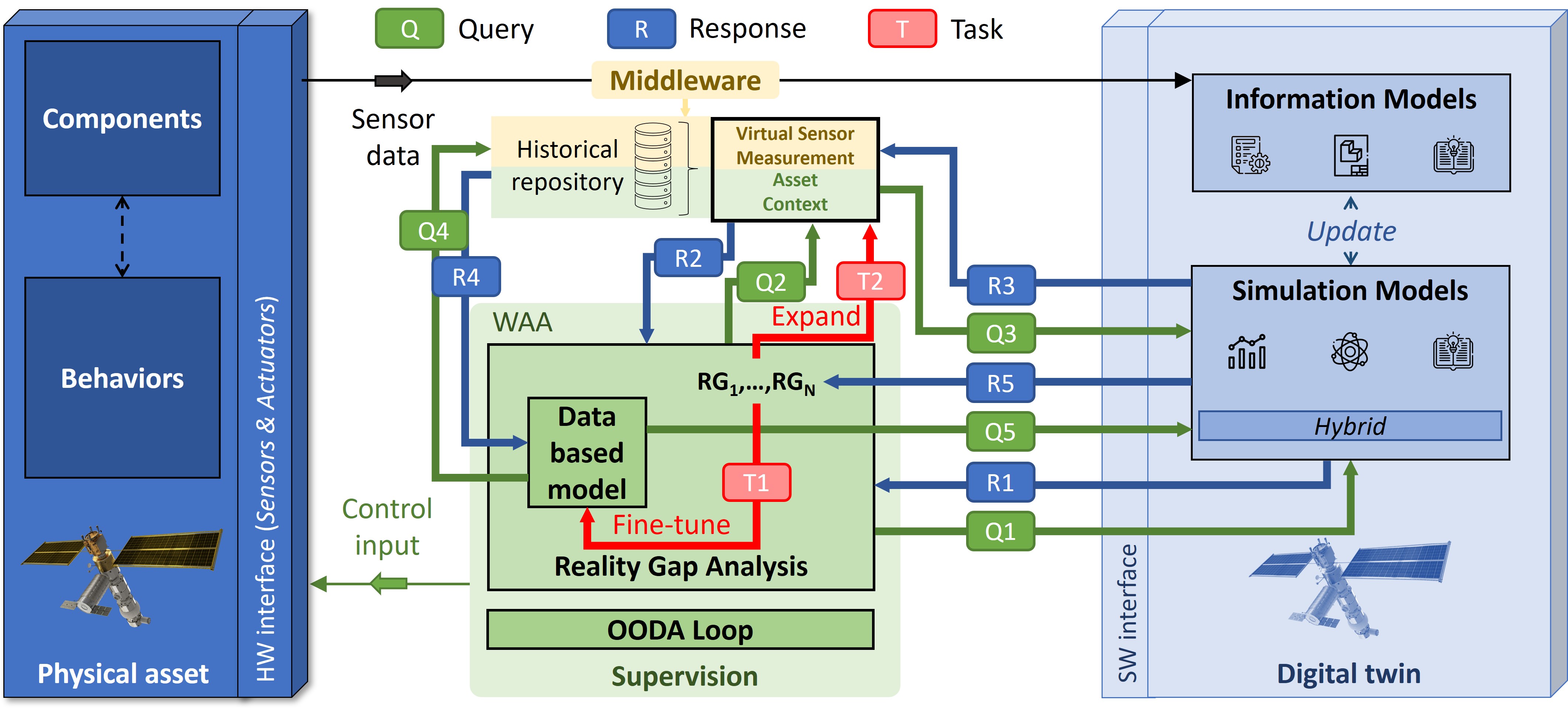}
    \caption{Proposed RGA module and its associated data pipeline.}
    \label{fig:framework_new}
\end{figure}

As discussed in Section~\ref{sec:lr}, a key challenge in DT calibration is ensuring structured information flow among system components so that context changes are captured in a timely and coherent manner. The RGA module addresses this challenge by focusing on three critical aspects: \vspace{-0.35cm}
\begin{enumerate}
    \item \textit{Role definition}: Clearly defining which entities (e.g., sensors, middleware, simulation engines, analytics modules) exchange information, under what conditions, and how these interactions contribute to maintaining DT calibration.
    \item \textit{Timing and synchronization}: Developing robust mechanisms to detect significant discrepancies between the digital twin's predictions and real-world measurements, triggering recalibration when necessary.
    \item \textit{Data integration}: Ensuring that newly acquired observations are incorporated into the DT (and corresponding simulation models) while preserving historical states and preventing conflicts in long-term data representation.
\end{enumerate} \vspace{-0.4cm}
To systematically address these challenges, we introduce the RGA module, which facilitates calibration through targeted information flow, out-of-sync detection, and dynamic data integration. Figure~\ref{fig:framework_new} outlines how these processes interact. Figure~\ref{fig:framework_new} highlights each query, response, and task involved in the process.  These elements---labeled as Q1-Q4 for queries, R1-R4 for responses, and T1-T2 for tasks---will serve as reference points throughout our discussion of the RGA in Section \ref{sec:method}.

At initialization, the RGA module queries the DT to verify that its simulation and information models contain sufficient detail to support accurate calibration (Q1 and R1). Specifically, it assesses the digital twin's ability to represent key physical parameters, sensor specifications, and contextual factors. If the DT lacks the necessary fidelity, the RGA pauses calibration and requests an update to the model’s parameters or structure. The RGA then checks the historical repository to determine whether relevant design-phase simulation data---previously generated synthetic sensor measurements tied to known contexts---are available (Q2 and R2). If such data are missing, the RGA requests new simulation runs (Q3 and R3) to generate baseline samples, ensuring that the simulation model can be compared against real-world measurements. Concurrently, the RGA initializes its data-driven model by querying the historical repository (Q4). This integrated dataset combines: (1) Virtual sensor measurements from the design phase or earlier operational phases where the context was explicitly known; and (2) real-world sensor measurements from live operations, where the system's underlying context was not directly measured.

Using this dataset, the data-driven model learns to map real-world sensor observations to latent contextual variables (R4). Once trained, it infers the system's current operating context from live sensor data. The inferred context is then communicated to the DT (Q5), enabling it to generate simulated sensor outputs (R5). Finally, the RGA then quantifies the reality gap by comparing real-world measurements \(\mathbf{y}_{t}^{\text{real}}\) to simulated outputs \(\mathbf{y}_{t}^{\text{sim}}\), providing a structured mechanism for detecting deviations between the DT and its physical assets.

Once the initial calibration is finished, the RGA module continuously monitors for deviations between the DT and its physical counterpart, identifying out-of-sync states that require recalibration (T1). At each time step \(t\), the RGA detects and address discrepancies: \vspace{-0.4cm}

\begin{enumerate}

\item \textit{Processing new sensor data}: Incoming real-world measurements, $\mathbf{y}_{t}^{\text{real}} = [y_{t,1}^{\text{real}}, \dots, y_{t,d}^{\text{real}}]$, are input into the data-driven model, which estimates the system's context, $\mathbf{c}_{t}$. This inferred context is used by the DT to generate corresponding simulated sensor outputs, $\mathbf{y}_{t}^{\text{sim}} = [y_{t,1}^{\text{sim}}, \dots, y_{t,d}^{\text{sim}}]$. By conditioning these predictions on the latest sensor data, the model captures changes in operational settings, reflecting more up-to-date real-world behavior.

\item \textit{Quantifying the reality gap}: The RGA computes the discrepancy between real and simulated measurements as a reality gap vector, \vspace{-0.6cm}
\[
\boldsymbol{\delta}_{t} \;=\; \Bigl[\delta_{t,1},\, \dots,\, \delta_{t,d}\Bigr], \vspace{-0.5cm}
\]
where each component $\delta_{t,i}$ measures the mismatch between real and simulated outputs for sensor $i$ using the squared error, \vspace{-0.6cm}
\[
\delta_{t,i} \;=\; \bigl(y_{t,i}^{\text{real}} - y_{t,i}^{\text{sim}}\bigr)^{2}, \vspace{-0.5cm}
\]

\item \textit{Detecting out-of-sync states and triggering recalibration}: The RGA maintains a sliding window of the most recent \(W\) reality gap values \(\{\delta_{t}, \delta_{t-1}, \ldots, \delta_{t-W+1}\}\), treating the error distribution for each sensor $i$ separately as a Gaussian distribution. Letting $\gamma_{t,i}$ denote the upper confidence bound for dimension~$i$ based on the recent window, the system classifies the DT as \textit{out-of-sync} when \(\delta_{t,i} \;>\; \gamma_{t,i}\) for any $i$. If an out-of-sync state is detected, the RGA initiates a recalibration cycle: (1) The latest real-world measurements are used to refine the data-driven model's parameters; (2) the updated model re-estimates \(\mathbf{c}_{t}\), improving its inference of the system's current context; and (3) the digital twin's corresponding simulation parameters are updated accordingly. This recalibration process ensures that the DT maintains predictive accuracy even as real-world operational conditions evolve. \vspace{-0.45cm}
\end{enumerate}

Ensuring that newly acquired observations are incorporated into the DT is critical for maintaining long-term accuracy.  The RGA addresses this through a structured integration process (T2), determining whether newly inferred context-virtual measurement pairs should be added to the historical repository. We define an upper confidence bound $\gamma'_{t,i}$ derived from the repository's historical distribution of errors for each sensor $i$. The RGA applies two key checks before storing new data: \vspace{-0.35cm}

\begin{enumerate}
\item \emph{Check the reality gap:} If: \vspace{-0.6cm}
\[
\delta_{t,i} \;\le\; \gamma'_{t,i}
\quad
\text{for all } i \,=\,1,\dots,d, \vspace{-0.5cm}
\]
this indicates that the reality gap vector, $\boldsymbol{\delta}_{t}$, falls within expected bounds across all sensor dimensions, making the new data reliable for future reference and model refinement.

\item \emph{Check context distinctness:} If the reality gap condition is satisfied, the RGA evaluates whether the inferred context, $\mathbf{c}_{t}$, represents a new system state. This is done by comparing $\mathbf{c}_{r}$ to all previously stored context vectors, $\mathbf{c}_{r}$, in the repository using a distance measure, $d(\mathbf{c}_{t}, \mathbf{c}_{r})$ (e.g., an $\ell_2$-norm). The new context is considered sufficiently distinct if: \vspace{-0.6cm}
\[
\min_{\mathbf{c}_{r}} d(\mathbf{c}_{t}, \mathbf{c}_{r}) > \tau_{\text{ctx}}, \vspace{-0.5cm}
\]
where $\tau_{\text{ctx}}$ is a predefined threshold for context distinctness. It determines whether the newly inferred context, $\mathbf{c}_{t}$, is sufficiently different from the contexts already stored in the repository. If this condition holds, the newly inferred context $\mathbf{c}_{t}$ and its corresponding simulated measurement $\mathbf{y}_{t}^{\text{sim}}$ are stored in the repository. If the gap condition is not met for at least one sensor dimension or if $\mathbf{c}_{t}$ fails to meet the distinctness criterion, the RGA withholds adding the new records. \vspace{-0.45cm}
\end{enumerate}

By requiring both a dimension-wise distribution check on the reality gap vector and a context-distinctness check, the repository expands only when new data fall within credible bounds and introduce novel information. Additionally, by explicitly defining communication pathways---specifying which entities interact, under what conditions, and how---and applying robust per-dimension checks to detect misalignment and manage data integration, the RGA ensures continuous context capture without relying solely on a single aggregated error metric. As a result, the DT maintains fidelity throughout the asset's lifecycle, adapting to evolving real-world conditions.

\subsection{Reality Gap Analysis for Context Learning and Adaptation} \label{sec:data_driven_model}

\begin{figure}[t]
    \centering
    \includegraphics[width=0.9\linewidth]{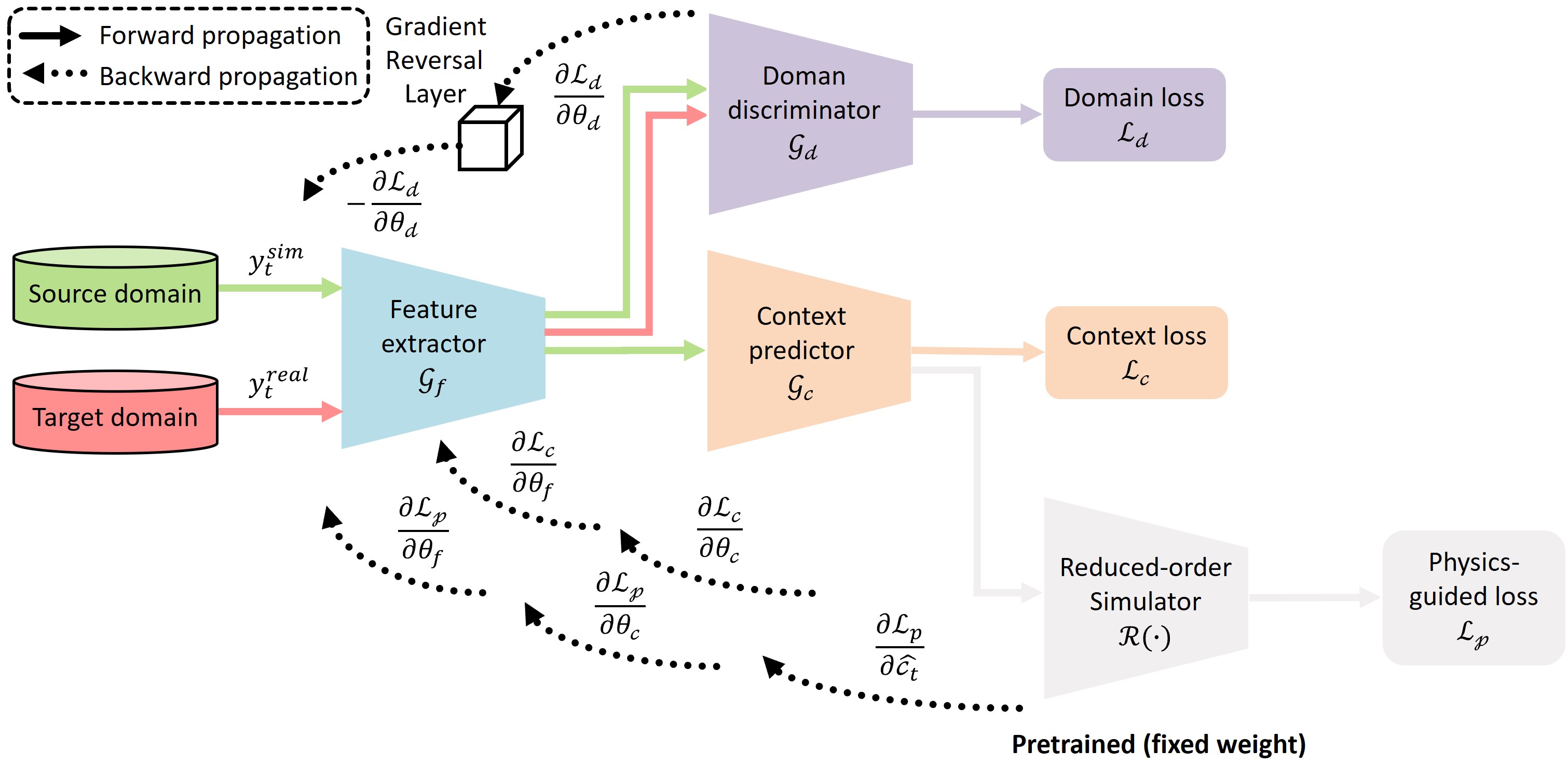}
    \caption{Architecture of the data-driven model used in the proposed RGA module, adapted from \protect\cite{ma_digital_2025}.}
    \label{fig:model}
\end{figure}

While the RGA module provides a structured approach for reality gap detection and calibration, its effectiveness depends on accurately inferring latent contextual variables from incoming sensor data. To achieve this, we integrate a deep learning-based approach that combines adversarial domain adaptation with physics-guided constraints. This section details the architecture of the data-driven model, which serves two primary objectives: (1) Extracting domain-invariant representations that align real-world (target domain) and simulated sensor data (source domain) distributions and (2) ensuring that inferred system states remain physically consistent through a reduced-order simulator. Figure~\ref{fig:model} provides a high-level overview of this architecture.

\subsubsection{Domain Adaptation for Real-world Alignment}

A challenge in using simulation data for real-world applications is the discrepancy between the two domains. Directly applying models trained in simulation to real-world sensor data can introduce significant errors.  To bridge this gap, we employ adversarial domain adaptation to enable the model to learn representations invariant to domain-specific differences. We first incorporate two common elements of adversarial domain adaptation. The \textit{encoder} (\(\mathcal{G}_{\text{f}}\) with weight \(\theta_{\text{f}}\) in Figure~\ref{fig:model}) maps sensor observations \(\mathbf{y}_{t}^{\text{real}}\) from the real-world domain or \(\mathbf{y}_{t}^{\text{sim}}\) from the simulation domain into a shared latent representation \(\mathbf{z}\). It extracts domain-invariant features, suppressing domain-specific characteristics so that downstream tasks can treat both sources uniformly. The \textit{domain discriminator} (\(\mathcal{G}_{\text{d}}\) with weight \(\theta_{\text{d}}\) in Figure~\ref{fig:model}) classifies whether a given latent embedding \(\mathbf{z}\) originates from the simulated domain (source) or from the real-world domain (target). During training, it provides a loss signal, \(\mathcal{L}_{\text{domain}}\) (\(\mathcal{L}_{\text{d}}\) in Figure~\ref{fig:model}), that drives the encoder to learn domain-invariant representations. Following standard practice \cite{ganin_domain-adversarial_2016}, the discriminator aims to maximize the accuracy of this classification, whereas the encoder---through adversarial training---attempts to minimize it.

These two components form the adversarial loop: as the discriminator becomes more proficient, the encoder adapts accordingly to produce features that the discriminator can no longer reliably distinguish. This approach has proven effective in sim-to-real transfer in robotics and DTs where simulation-trained models must adapt to real-world data \cite{zhao2020transfer}.

\subsubsection{Context Inference via Deep Learning}
Beyond aligning domain representations, the model must infer context variables, \(\mathbf{c}_{t}\), from incoming sensor data. Context variables encode critical environmental and operational factors that influence the behavior of the physical asset and its DT. To enable accurate context inference, we introduce a \textit{context predictor} (\(\mathcal{G}_{\text{c}}\) with weight \(\theta_{\text{c}}\) in Figure~\ref{fig:model}) that builds on the encoder's output. The context predictor operates as a decoder that takes the shared latent representation, \(\mathbf{z}\), from the encoder as input and produces an estimate of the current system context \(\hat{\mathbf{c}}_{t}\). 

During inference, once the encoder and context predictor are initialized or fine-tuned, they operate together to transform sensor observations into the latent context needed for calibration. The corresponding loss, \(\mathcal{L}_{\text{context}}\) (\(\mathcal{L}_{\text{c}}\) in Figure~\ref{fig:model}), measures how closely \(\hat{\mathbf{c}}_{t}\) matches the true or expected context (for simulation data where context is known) and provides another training signal that updates the encoder. This is essential for bridging observations with the variables (e.g., operating modes, boundary conditions) that the DT requires for more accurate simulation.

\subsubsection{Enforcing Physical Constraints with a Reduced-order Simulator}
While adversarial domain adaptation aligns real-world and simulated data distributions, it does not guarantee that inferred contexts, \(\hat{\mathbf{c}}_{t}\), obey the underlying physical laws governing the system. Without explicit enforcement of domain-specific constraints, context predictions may drift toward statistically plausible but physically meaningless states. To ensure that inferred contexts remain physically valid, we introduce a \textit{reduced-order simulator} into the learning process as an additional neural module. This simulator provides an additional layer of physical consistency by mapping inferred contexts to virtual sensor measurements: \vspace{-0.3cm}
\begin{enumerate}
    \item \textit{Pre-trained reduced-order simulator}:
    This network approximates the underlying physics---such as partial differential equations or mechanistic models---while providing computationally efficient and differentiable insights. We treat it as a ``black-box'' function, \(\mathcal{R}(\cdot)\), that maps a context estimate \(\hat{\mathbf{c}}_{t}\) to a corresponding virtual measurement \(\hat{\mathbf{y}}_{t}^{\text{sim}}\). 
    \item \textit{Physics-guided loss}:
    During training, we compare \(\hat{\mathbf{y}}_{t}^{\text{sim}} = \mathcal{R}\bigl(\hat{\mathbf{c}}_{t}\bigr)\) with the actual sensor observation \(\mathbf{y}_{t}^{\text{real}}\). The discrepancy between these two values forms a physics-guided loss term, \(\mathcal{L}_{\text{physics}}\) (\(\mathcal{L}_{\text{p}}\) in Figure~\ref{fig:model}), which pushes \(\hat{\mathbf{c}}_{t}\) to align not just with the distribution of contexts but also with physically consistent outcomes, thereby penalizing predictions that fail to reproduce the physical signals observed in the real system.
\end{enumerate}
\vspace{-0.3cm} Once trained, the reduced-order simulator is kept fixed in its weights during the subsequent initializing and fine-tuning stages.  This prevents domain adaptation updates from distorting fundamental physical relationships, maintaining a stable representation of system behavior.

\subsubsection{Training and Deployment for Continuous Calibration}
We combine the three loss components---\(\mathcal{L}_{\text{domain}}\), \(\mathcal{L}_{\text{context}}\), and \(\mathcal{L}_{\text{physics}}\)---during both initialization and fine-tuning stages: \vspace{-0.6cm}
\[
\mathcal{L}_{\text{total}} \;=\; \alpha\,\mathcal{L}_{\text{domain}} \;+\; \beta\,\mathcal{L}_{\text{context}} \;+\; \gamma\,\mathcal{L}_{\text{physics}}, \vspace{-0.5cm}
\]
where \(\alpha\), \(\beta\), and \(\gamma\) are weighting coefficients that balance the importance of domain invariance, context prediction accuracy, and physical fidelity. Once training or fine-tuning converges, only the encoder and context predictor are actively used to infer \(\mathbf{c}_{t}\) in real-time operations, while the other components (i.e., domain discriminator and reduced-order simulator) can remain offline or operate at lower frequency depending on the calibration schedule.

The primary advantage of this design lies in its ability to leverage domain adaptation techniques for bridging simulation and real-world data, while also incorporating physics knowledge through a lightweight, pre-trained simulator. By introducing a reduced-order simulator in the training loop and comparing its output to actual sensor measurements, we ensure that the inferred context is rooted in feasible physical behavior. This approach reduces the risk of learning spurious or unphysical mappings. Furthermore, by fixing the simulator's weights, we retain a stable embedding of the governing physics and prevent these relationships from drifting during adversarial training. 

Beyond inverse modeling considerations, this design also confers other benefits: it alleviates the need for full-scale physics solvers by using the simulator’s compact representation, thus enhancing computational efficiency for real-time deployment. Coupled with adversarial domain adaptation, the model becomes robust to domain shifts between simulation and real-world distributions, enabling effective, continuous calibration of the DT under evolving operating conditions.

\section{Illustrative Case Study and Benchmarking} \label{sec:case}
\subsection{Case Study Overview} \label{sec:case_overview}
To demonstrate the effectiveness of the proposed RGA module, we present an illustrative case study applying our data-driven method within the DT framework. Specifically, we focus on the Newell-Simon Bridge, a pedestrian steel truss bridge located on Carnegie Mellon University's campus (Pittsburgh, PA). This structure---shown in Figure~\ref{fig:NSB}---serves as a representative asset for condition-based monitoring, allowing us to better understand how our proposed RGA module and the associated information flows detailed in the previous section could help maintain alignment between real-world sensor data and a simulation model over time. 

\subsubsection{DT Setup for Newell-Simon Bridge Monitoring}

\begin{figure}
\centering
\begin{subfigure}[b]{0.35\textwidth}
   \includegraphics[width=1\linewidth]{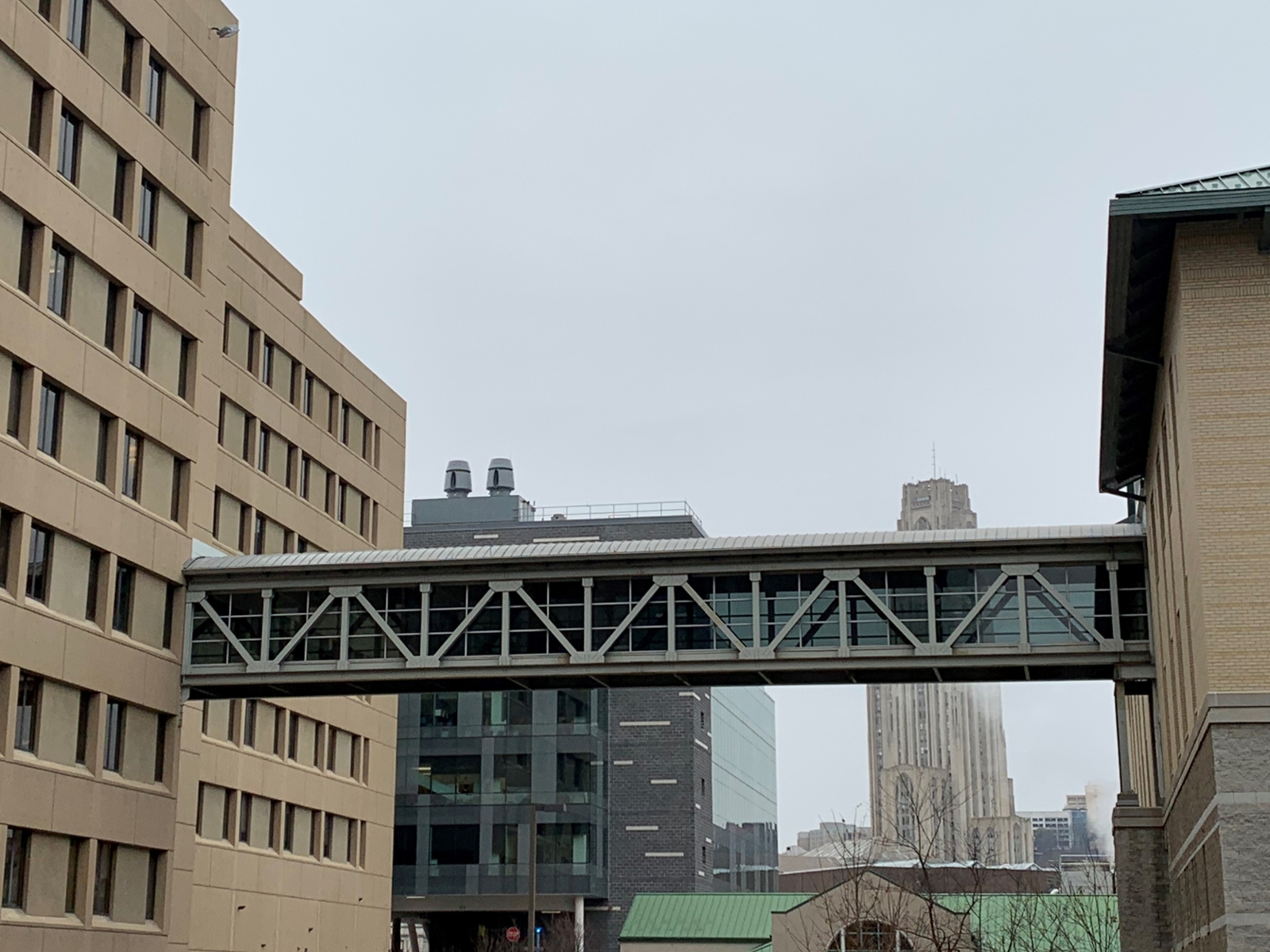}
   \caption{}
   \label{fig:NSB} 
\end{subfigure}
\hfill
\begin{subfigure}[b]{0.58\textwidth}
   \includegraphics[width=1\linewidth]{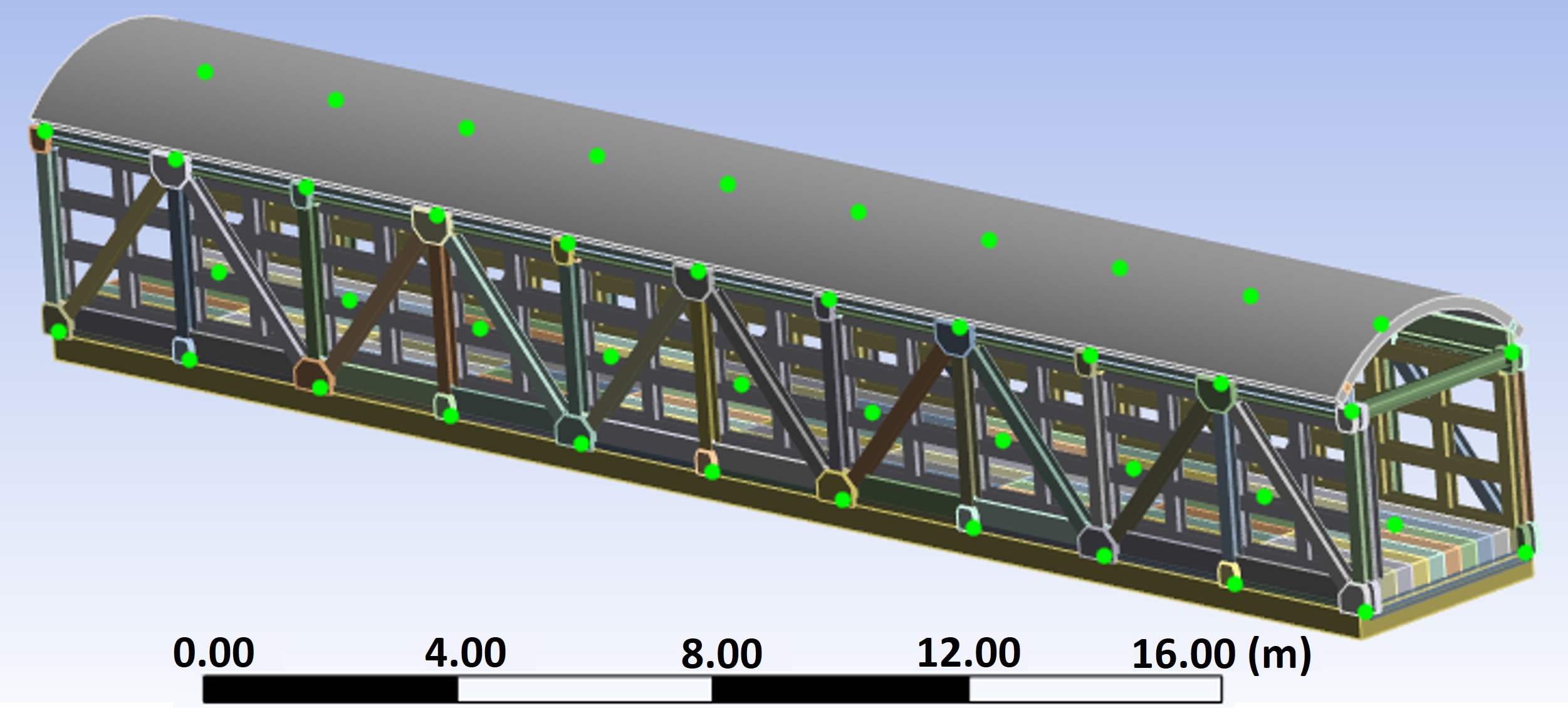}
   \caption{}
   \label{fig:NSB-Ansys}
\end{subfigure}

\caption{Newell-Simon Bridge's (a) physical twin and (b) DT with virtual sensors highlighted in green.}
\end{figure}

In this proof-of-concept, our physical twin is implemented as a high-fidelity finite element model (FEM) of the Newell-Simon Bridge. This simulation model features 42 virtual sensors on each outer truss to measure structural deformations (highlighted in green in Figure~\ref{fig:NSB-Ansys}). Artificial noise is introduced to emulate the variability found in real-world signals. We use Ansys Mechanical for structural simulations, while Python automates the integration of sensor feedback, enabling efficient synchronization between the simulated ``physical'' phenomena and the DT.

\subsubsection{Levels of Integration for Evaluating the RGA}

\begin{figure*}
    \centering
    \includegraphics[width=0.9\linewidth]{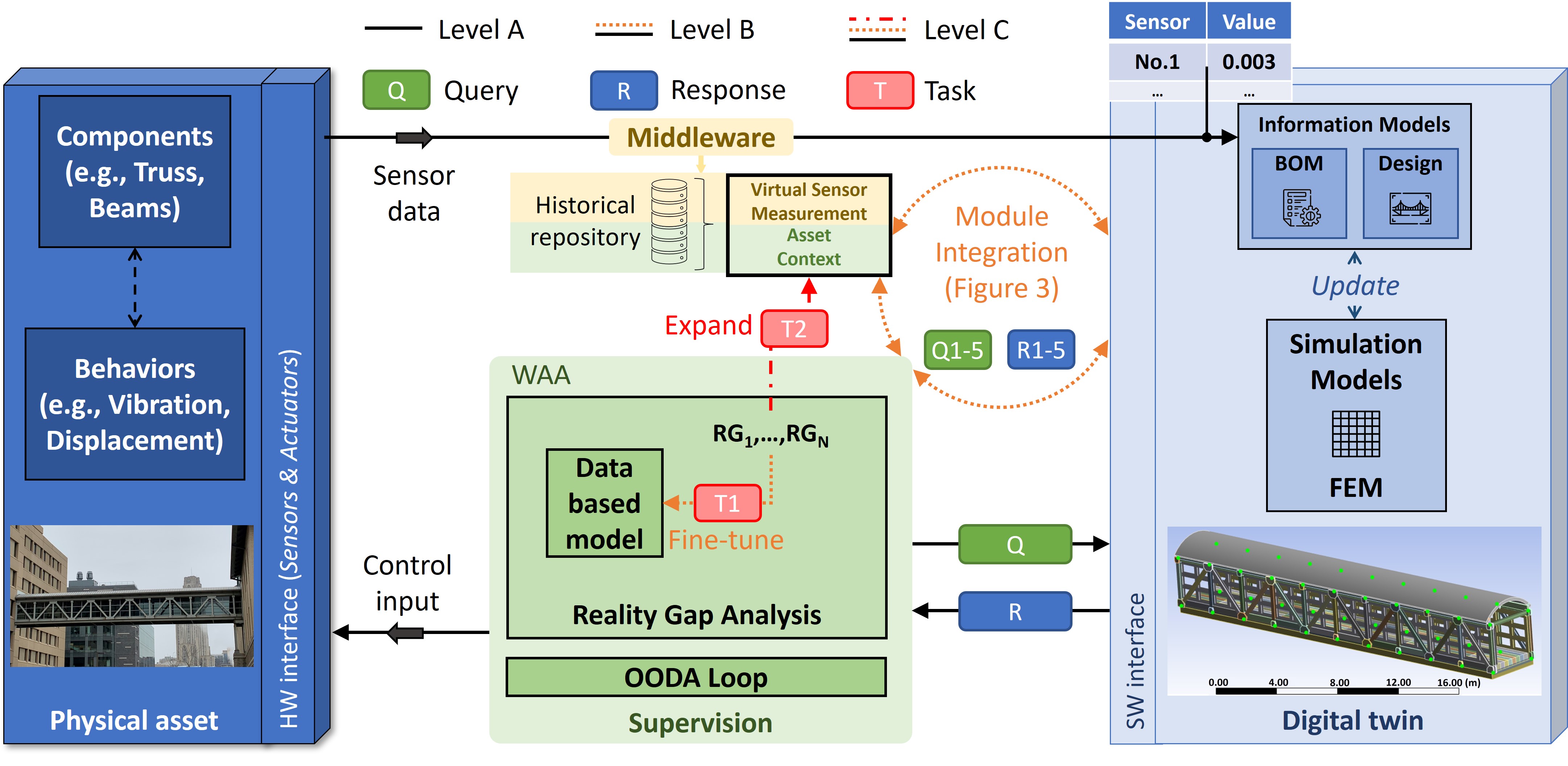}
    \caption{Step-by-step implementation of the case study.}
    \label{fig:LoI}
\end{figure*}

To systematically evaluate the incremental benefits of each core feature of the RGA module, we define three Levels of Integration (LoI), as shown in Figure~\ref{fig:LoI}. \emph{LoI~A} implements only an initial domain-adversarial training phase to calibrate the DT without subsequent recalibration; \emph{LoI~B} extends LoI~A by adding the detection of out-of-sync states for recalibrating the DT (T1) whenever the reality gap grows significantly; and \emph{LoI~C} incorporates both detection of out-of-sync states and a repository expansion mechanism (T2) to integrate newly discovered novel scenarios---that lie within credible bounds---into the digital twin’s historical library. This tiered evaluation structure highlights how each additional feature---continuous adaptation and real-time expansion---contributes to reducing the reality gap under real-world constraints.

\subsubsection{Data Partitioning and Experimental Design}

We partition the simulation data into three subsets: 50\% for training, 20\% for validation, and 30\% for testing. Unlike conventional train-test splits, this design designates the validation set to emulate unlabeled real-world data arriving incrementally during deployment, including potential distribution shifts. In line with our initialization step from LoI~A., the training set (50\%) initializes the domain-adversarial network on synthetic-labeled plus minimal real-unlabeled samples. Next, once the model has been trained, the validation set (20\%) acts as a proxy for online adaptation. We deliberately introduce artificial drifts partway through these validation samples (e.g., modifying sensor noise or operating conditions), triggering our detection of out-of-sync states for recalibrating the DT (Section~\ref{sec:gap1}). Under LoI~B or LoI~C, the RGA module then fine-tunes the data-driven model on newly arrived validation data, either performing recalibration of the DT alone (LoI~B) or also expanding the historical repository with novel configurations (LoI~C). By positioning the recalibration in the validation stream, we ensure that adaptation is tested on a near-realistic flow of incoming sensor measurements, without contaminating the final evaluation data. Finally, the test set (30\%) is used only for performance checks after the calibration or recalibration steps for the DT.

\subsubsection{Evaluation Metrics}

To compare the effectiveness of each LoI, we define three key performance metrics and their associated units (or lack thereof). First, we measure the accuracy of predicted context (``Error'' in Table~\ref{tab:LoIcomparison}) during testing by computing a weighted mean squared error (MSE) of the inferred context vector. Because each context variable has a different scale and physical unit (e.g., temperature in °C, load in kN, etc.), we normalize them before computing the weighted MSE to obtain a dimensionless error metric, where each variable’s weight is determined by its range observed in the simulation-derived context data. Second, we evaluate the mean reality gap (``RG'') across individual sensors over the entire test set. Specifically, we compare real and simulated sensor measurements (in millimeters) and then take the average of their squared differences, yielding an RG value in \(\text{mm}^2\) for our structural displacement application. Finally, we track the average delay (``AD'') in detecting drifts that trigger recalibration, which is a unitless count of the number of samples required before the out-of-sync state is identified. In practical deployments, this sample-based AD can be readily translated into time by multiplying by the data collection frequency (e.g., an AD of 10 corresponds to 10 seconds at a 1\,Hz sampling rate, excluding additional hardware or network delays).

\begin{table} \small
    \centering
\begin{tabular}{cccccccccc}
  \toprule
  \multirow{2}{*}{LoI}& \multicolumn{3}{c}{\begin{tabular}{@{}c@{}}\textbf{With residual loss} \\ \cite{i3ce2025}\end{tabular}} & \multicolumn{3}{c}{\textbf{Without physics-guided loss}} & \multicolumn{3}{c}{\textbf{With physics-guided loss}}  \\  
   & Error [/] & RG [\(\text{mm}^2\)] & AD [/] & Error [/] & RG [\(\text{mm}^2\)] & AD [/] & Error [/] & RG [\(\text{mm}^2\)] & AD [/] \\ \toprule
  \multirow{1}{*}{LoI A} & 0.466 & 352 & N/A$^\dagger$ & 0.681 & 512 & N/A$^\dagger$ & 0.405 & 297 & N/A$^\dagger$
   \\             \cmidrule(lr){2-4} \cmidrule(lr){5-7} \cmidrule(lr){8-10}

    \multirow{1}{*}{LoI B} & 0.387 & 301 & 56.8 & 0.674 & 520 & 70.2 & 0.376 & 259 & 49.4
   \\             \cmidrule(lr){2-4} \cmidrule(lr){5-7} \cmidrule(lr){8-10}

   \multirow{1}{*}{LoI C} & 0.392 & 298 & 57.1 & 0.672 & 526 & 69.8 & 0.379 & 261 & 50.3
   \\             \bottomrule

\end{tabular}

\begin{tablenotes}
      \footnotesize
      \item  \hspace{0.35cm} $^{\dagger}$Re-calibration mechanism is not implemented for LoI A.
    \end{tablenotes}

    \caption{Performance compared across different LoIs averaged on 10 random splits. }
    \label{tab:LoIcomparison}
\end{table}

\subsection{Results and Discussion} \label{sec:results_discussion}

Table~\ref{tab:LoIcomparison} summarizes each LoI's performance under three conditions: (1) A prior residual-based neural network approach~\cite{i3ce2025}; (2) a domain-adversarial training approach without physics-guided loss; and (3) the same domain-adversarial approach with physics-guided loss. 

\subsubsection{Performance of LoI~A}
LoI~A implements only the initial domain-adversarial training phase, similar to a conventional ``train-and-deploy'' paradigm without subsequent recalibration or repository updates. As shown in Table~\ref{tab:LoIcomparison}, LoI~A already achieves better context inference and a smaller reality gap (RG) than the residual-based method~\cite{i3ce2025} when augmented with physics-guided loss. This improvement highlights the benefits of incorporating physics-informed constraints during network training: by aligning learned representations with domain-specific knowledge, the model reduces the mismatch between real and simulated signals. However, LoI~A's static nature cannot correct emerging sensor drift or unexpected environmental changes that arise after deployment. Once the model is trained, it lacks a feedback loop to update its parameters, which makes it prone to accumulation of systematic errors. For instance, if temperature variations or material fatigue alter the dynamics of the Newell-Simon Bridge over time, LoI~A will not capture these shifts. As a result, the reality gap may widen in real-world scenarios that deviate significantly from the initial training distribution, which further manifests into inaccurate what-if analysis or maintenance scheduling.

\subsubsection{Performance of LoI~B}
LoI~B addresses the limitations of LoI~A by introducing a mechanism to detect out-of-sync states, fine-tune the data-driven model, and recalibrate the DT (T1). When the RGA module flags persistent discrepancies in the validation stream, LoI~B updates its encoder, context predictor, and domain-classifier modules to better align the digital twin's simulations with new operating regimes. Comparing LoI~B to LoI~A in Table~\ref{tab:LoIcomparison} reveals several notable trends. First, LoI~B achieves further reductions in RG and improves context inference, demonstrating the value of ongoing recalibration when faced with previously unseen conditions. Second, the average delay (AD) in detecting out-of-sync behavior declines, especially when physics-guided losses are included. This improvement can be attributed to the structural consistency enforced by the physics-based constraints: because the model’s outputs are grounded in realistic physical relationships, any drift in sensor readings is more readily identified as an anomaly. In other words, the digital twin’s baseline predictions remain coherent enough that deviations caused by new environmental conditions or sensor degradation become easier to detect earlier. Consequently, the RGA module can trigger fine-tuning promptly, ensuring that the evolving physical system is mirrored more accurately by the DT. 

\subsubsection{Performance of LoI~C}
LoI~C extends LoI~B by incorporating a confidence-based repository expansion (T2), which stores newly discovered operational regimes deemed credible and sufficiently distinct from previously encountered scenarios. This “continuous learning loop” grants the DT an opportunity to accumulate knowledge about conditions that fall outside its original training domain. Table~\ref{tab:LoIcomparison} shows that LoI~C often delivers modest near-term improvements over LoI~B in terms of accuracy and RG. However, the more significant advantage emerges when similar ``novel'' conditions reoccur later. Because the DT has already integrated those contexts into its historical library, it can rapidly adjust its predictions without large-scale retraining. This leads to quicker drift detection and less pronounced misalignments over repeated exposures to similar changes. Although these benefits may appear incremental in short-duration tests, the long-term payoff is substantial. As the repository grows over extended periods, the model is better prepared for a broader range of operational scenarios, from seasonal temperature swings to gradual structural aging. In effect, LoI~C creates a self-reinforcing adaptation mechanism: each time the system confronts a new context, the expanded repository enhances future simulations and expedites subsequent recalibrations. This characteristic is especially valuable for large-scale or critical-infrastructure applications, where the cost of persistent mismatches or delayed re-calibration can be high. 

\subsubsection{Summary and Implications}
The results confirm that physics-guided, adversarial domain adaptation---augmented with continuous recalibration and repository expansion---supports more accurate, robust, and responsive DT calibration and recalibration compared to purely data-driven or residual-based baselines. While LoI~A demonstrates the core utility of physics-informed adversarial learning for digital twins, its inability to adapt post-deployment underscores the need for continuous feedback loops. LoI~B addresses emerging discrepancies with on-the-fly retraining, reducing detection delays when drifts occur. Finally, LoI~C complements recalibration with systematic repository growth, further future-proofing the DT against recurrent or cumulative operational shifts.  These findings highlight how the proposed RGA module, combined with iterative model refinement and physics-guided constraints, can better preserve alignment between a virtual model and its real-world counterpart over time. 

\section{Conclusion} \label{sec:conclusion} 
This paper introduces an RGA module designed to continuously reduce context mismatch between DTs and their physical counterparts. The module emphasizes calibrating the DT by capturing and updating latent real-world context, using knowledge transferred from an explicit and comprehensive simulation context. Unlike conventional sim-to-real approaches that often perform a one-off policy transfer, our work positions context inference and recalibration as ongoing processes throughout an asset’s lifecycle. To accomplish this, we develop a DT architecture that separates real-world data acquisition, simulation model updates, and context inference through an intermediary query-and-response mechanism. At its core, the RGA uses a deep learning model trained with adversarial domain adaptation and guided by a physics-based loss function derived from a reduced-order simulator. This design enforces domain-invariant feature learning while preserving physical consistency. Through a case study involving a steel truss bridge, the RGA module demonstrated its ability to rapidly detect out-of-sync states and improve calibration accuracy. By benchmarking multiple LoIs, we highlight how each additional feature---from out-of-sync detection to repository expansion---helps maintain robust alignment between the digital and physical counterparts.

While our approach focuses on deep learning-based context inference, a wide range of techniques could be used to reduce context mismatch and close the reality gap in DTs. For example, Bayesian updating methods, ensemble filtering approaches, and rule-based expert systems may also offer advantages in domains where data are scarce or physical laws are more explicitly formulated. When closed-form models of the physics are tractable, physics-informed neural networks, robust state observers, or model-driven Kalman filters could also be viable alternatives for continuous calibration. More broadly, a comprehensive benchmarking of these methods---encompassing both data-driven and model-based paradigms---is needed to establish clearer guidelines on their respective strengths and limitations under various operational conditions. Such experiments will require carefully designed protocols and cross-validation strategies to ensure fair comparability across algorithms. Although our results underscore the promise of adversarial domain adaptation with physics-guided constraints, future research should explore how different approaches can be combined, tailored, or selected to further narrow the reality gap in diverse DT applications.

\section{Data Availability Statement}
Some or all data, models, or code that support the findings of this study are available from the corresponding author upon reasonable request.



\bibliography{ref}

\end{document}